\documentclass[twoside,11pt]{article}
\usepackage{amsfonts,amssymb,latexsym,amscd}
\usepackage{amsmath}
\usepackage{soul} 
\usepackage{bm}
\usepackage{mathtools}
\usepackage[utf8]{inputenc}
\usepackage[english]{babel}
\usepackage{booktabs}
\usepackage{graphicx}
\usepackage{tikz}
\usepackage{epstopdf}
\usepackage{subfig}
\usepackage{setspace}

\usepackage{tikz}
\usetikzlibrary{calc}
\usetikzlibrary{shapes,fit}

\usetikzlibrary{automata}
\usetikzlibrary{patterns}
\usetikzlibrary{fit,positioning}
\pgfmathsetmacro{\sinOffset}{sin(45)}
\pgfmathsetmacro{\cosOffset}{cos(90)}
\newcommand{\bulletsize}{9}
\newcommand{\nodedist}{4}
\tikzstyle{cont}=[circle,accepting, minimum size = \bulletsize mm, fill = white!100, thick, draw =black!80, node distance = \nodedist mm] 
\tikzstyle{disc}=[circle, minimum size = \bulletsize mm, fill = white!100, thick, draw =black!80, node distance = \nodedist mm]
\tikzstyle{obsv}=[circle, minimum size = \bulletsize mm, fill = blue!15, thick, draw =black!80, node distance =\nodedist mm]
\tikzstyle{marg}=[pattern=north west lines, pattern color=red!40]
\tikzstyle{parm}=[circle, minimum size = \bulletsize mm, node distance =\nodedist mm]
\tikzstyle{dist}=[node distance =\nodedist mm]
\tikzstyle{dummy}=[circle, minimum size = \bulletsize mm, node distance =\nodedist mm]
\tikzstyle{connect}=[-latex, thick,>=latex]
\tikzstyle{line}=[thick]
\tikzstyle{shaded}=[draw=black!20,fill=black!20]
\tikzstyle{box}=[rectangle, inner sep=2.5mm,draw=black!100, thick,dashed,rounded corners]
\tikzstyle{boxblue}=[rectangle, inner sep=2mm,fill=blue!20, thick,dashed,rounded corners]

\begin{document}
\title{The Nataf-Beta Random Field Classifier: An Extension of the Beta Conjugate Prior  to Classification Problems}

\author{\textbf{James-A. Goulet}\footnote{e-mail: james.a.goulet@gmail.com}
\mbox{}\\ 
\small Department of Civil and Environmental Engineering\\
\small \textsc{University of California, Berkeley}\\
\small Berkeley, \underline{USA}\\}
\maketitle
\begin{abstract}
This paper presents the \emph{Nataf-Beta Random Field Classifier}, a discriminative approach that extends the applicability of the Beta conjugate prior to classification problems. The approach's key feature is to model the probability of a class conditional on attribute values as a random field whose marginals are Beta distributed, and where the parameters of marginals are themselves described by random fields. Although the classification accuracy of the approach proposed does not statistically outperform the best accuracies reported in the literature, it ranks among the top tier for the six benchmark datasets tested. The \emph{Nataf-Beta Random Field Classifier} is suited as a general purpose classification approach for real-continuous and real-integer attribute value problems.
\end{abstract}

\section{Introduction}\label{S:Intro}
A large number of classification algorithms have been developed and are already achieving excellent performances in real-life classification contexts \cite{murphy2012machine}. The goal of this paper is to present a new method that extends the applicability of the \emph{Beta conjugate prior} to classification problems. The main incentives for such a method are (1) to have a probabilistic framework that is genuinely compatible with classification problems, and (2) to allow for the same intuitive interpretation as the Beta conjugate prior where the posterior probability density function (\emph{pdf}) describing the probability of a class depends on the number of the number of positive and negative observations. 

This paper presents a new discriminative classification approach; its key feature is to model the probability of a class conditional on attribute values as a random field whose marginals are Beta distributed, and where the parameters of marginals are themselves described by random fields. Section \ref{S:methodology} presents the mathematical formulation for the \emph{Nataf-Beta Random Field Classifier}; Section \ref{S:validation} validates the approach using both simulated and benchmark datasets; Section \ref{S:comparison} compares the approach proposed with Gaussian Process classification, a methodology that also relies on random fields; Section \ref{S:Discussion} discusses the limitations of the current approach and provides guidance for future extensions. 

\section{Methodology}\label{S:methodology}
The notation used in this paper is the following: Lower-case letters, e.g. ``$x$'' denote standard variables and indexes; Upper-case letters,``$X$'', denote random variables. A hat symbol denotes an estimation, e.g. ``$\hat{x}$''. Bold characters, i.e. ``$\mathbf{x} \text{ or }\mathbf{X}$'' represent matrices and vectors, and calligraphic letters ``$\mathcal{X}$'' represent sets. Lower-case Courier fonts represent length of sets and vectors, e.g. $\mathbf{x}=[x_{1},x_{2},\cdots,x_{\mathtt{x}}]$. $f(x)=\Pr(X=x)$ denotes a probability density function (\emph{pdf}). $F(x)=\Pr(X\leq x)$ denotes a cumulative density function (\emph{cdf}). Superscripts $f'(x)$ and $f''(x)$ respectively denotes prior and posterior \emph{pdf}. The tilde symbol, i.e., $\tilde{f}(x)$ denotes the predictive estimate of $f(x)$.

\begin{figure}[htbp]
	\begin{center}
	\includegraphics[width=85mm]{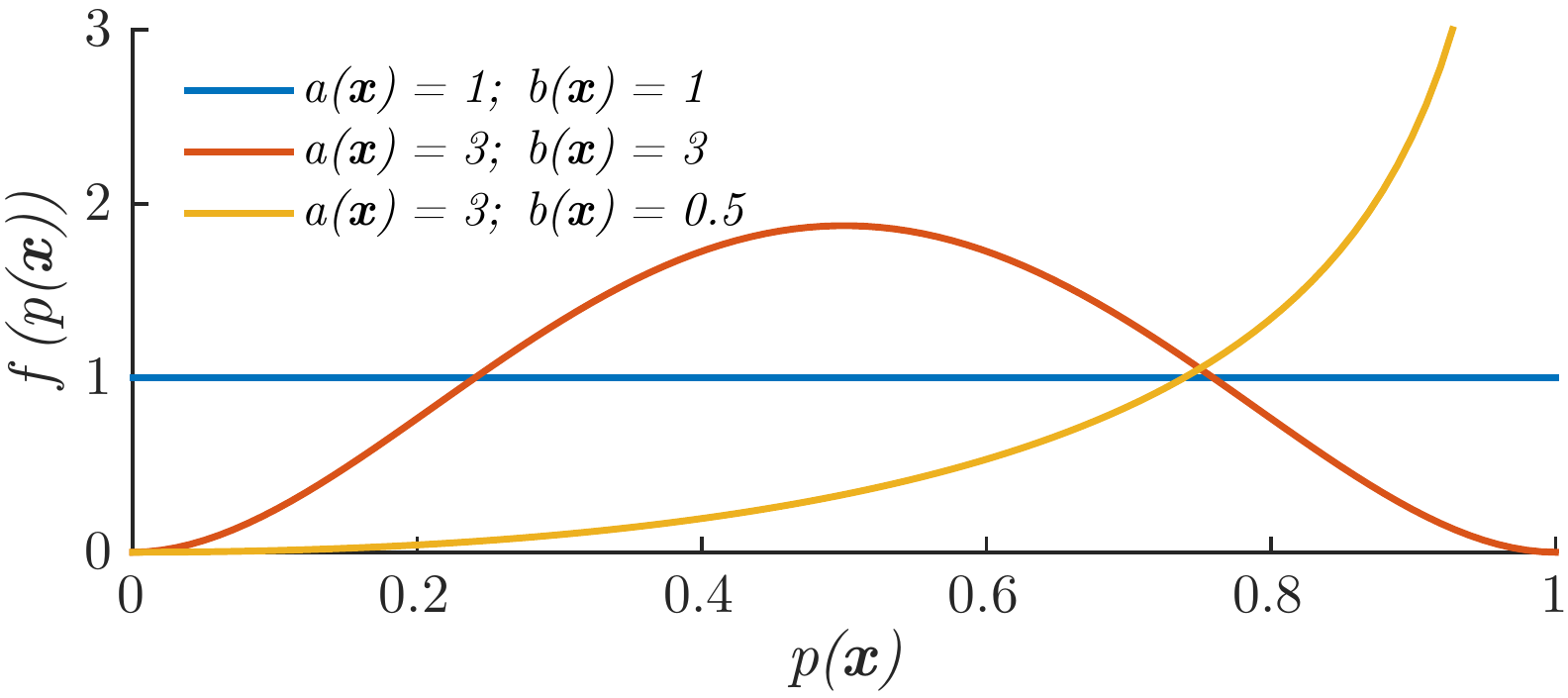}
	\caption{Examples of Beta \emph{pdf}s describing the posterior knowledge of $p(\mathbf{x})$, given sets of observations.}
	\label{fig:beta_pdf}
	\end{center}
\end{figure}

\begin{figure}[htbp]
	\begin{center}
	\includegraphics[width=80mm]{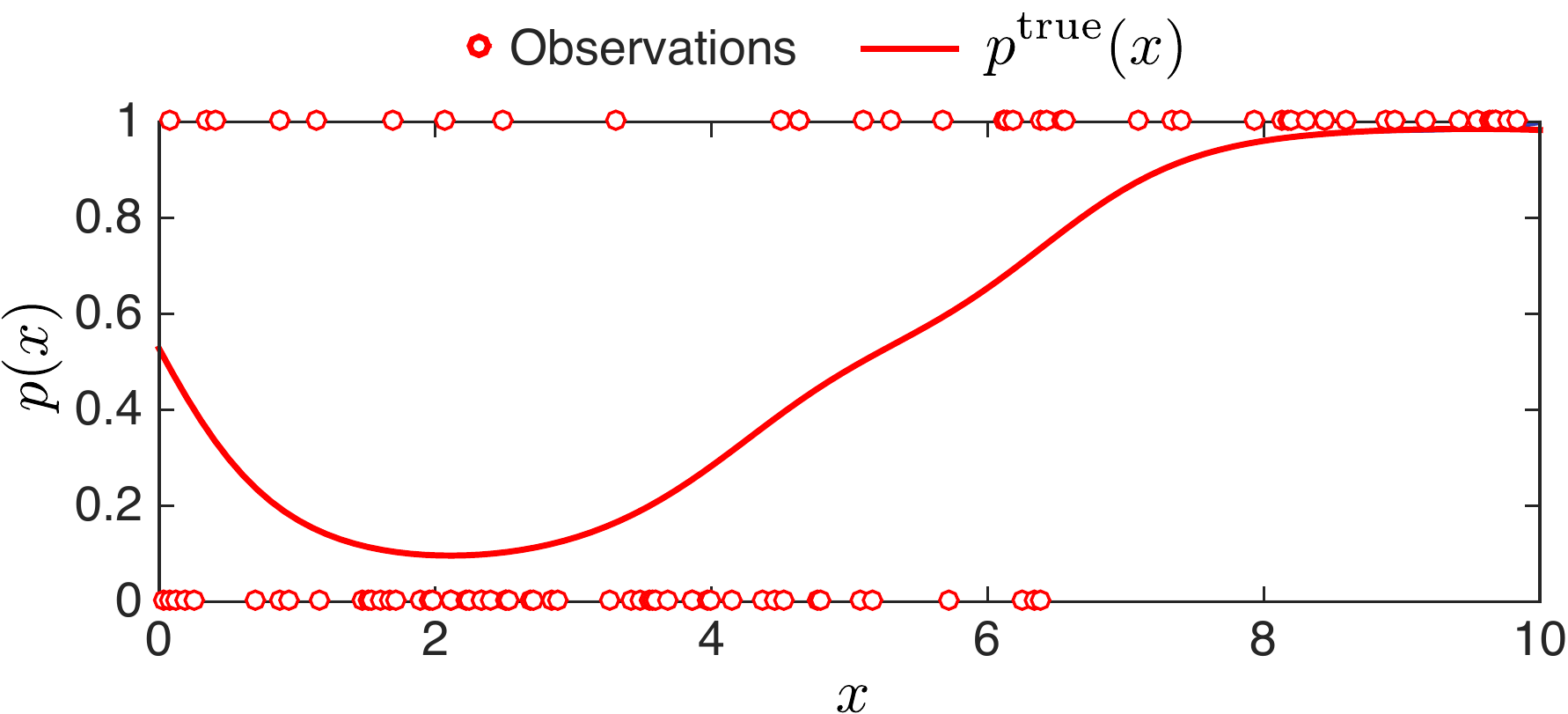}
	\caption{Unidimensional example of $p^{\text{true}}(x)$ where dots corresponds to $\mathtt{d}$ class observations, each corresponding to a different attribute value $x$.}
	\label{fig:Simulated_data}
	\end{center}
\end{figure}

Given $c\in\{0,1\}$ a binary indicator variable referring to either class ``0'' or ``1'',  $c(\mathbf{x})$ describes a class label as a function of a $\mathtt{x}$-dimensionnal vector containing continuous attributes values, $\mathbf{x}=[x_{1},x_{2},\cdots,x_{\mathtt{x}}]^{\intercal}$, where for all $i, x_{i}\in \mathbb{R}$. The knowledge of the possible values of $c(\mathbf{x})$ is represented by a Bernouilli random variable $C(\mathbf{x})\sim\text{Ber}(p(\mathbf{x}))$, where $p(\mathbf{x})\triangleq\Pr(C(\mathbf{x})=1)$. Given a set of statistically independent observations $\mathcal{D}=\{\hat{c}_{i}(\mathbf{x}_{i})\}_{i=1}^{\mathtt{d}}$ all sharing a common vector of attribute values $\mathbf{x}_{i}=\mathbf{x}^{*},\forall i$, the posterior \emph{pdf} of $p(\mathbf{x})$ can be defined using the Beta conjugate prior \cite{murphy2012machine}, so that
\begin{equation}
f''\left(p(\mathbf{x}^{*})|\mathcal{D}\right)\triangleq\text{Beta}(p(\mathbf{x}^{*});a(\mathbf{x}^{*}),b(\mathbf{x}^{*}))=\frac{p(\mathbf{x}^{*})^{a(\mathbf{x}^{*})-1}(1-p(\mathbf{x}^{*}))^{b(\mathbf{x}^{*})-1}}{\text{B}(a(\mathbf{x}^{*}),b(\mathbf{x}^{*}))}
\label{EQ:Beta_post}
\end{equation}
$a(\mathbf{x}^{*})$ and $b(\mathbf{x}^{*})$ respectively corresponds to the number of observations $\hat{a}(\mathbf{x}^{*})$ and $\hat{b}(\mathbf{x}^{*})$, where $\hat{c}(\mathbf{x}^{*})=1$ or $0$ so that
\begin{equation}
\begin{array}{rcl}
a(\mathbf{x}^{*})=\hat{a}(\mathbf{x}^{*})&\triangleq&\#\{i:\{ \hat{c}_{i}(\mathbf{x}^{*})=1\}_{i=1}^{\mathtt{d}}\}\\[4pt]
b(\mathbf{x}^{*})=\hat{b}(\mathbf{x}^{*})&\triangleq&\#\{i:\{ \hat{c}_{i}(\mathbf{x}^{*})=0\}_{i=1}^{\mathtt{d}}\}
\end{array}
\label{EQ:ab_counts}
\end{equation}
Figure \ref{fig:beta_pdf} presents examples of Beta \emph{pdf}s for several sets of observations. 
If an infinite number of observations are available for all $\mathbf{x}$, the approach above will provide accurate estimates of the true probability $p^{\text{true}}(\mathbf{x})$ so that 
$$
\underset{\mathtt{d}\to \infty}{\lim} f''(p(\mathbf{x})|\mathcal{D})=\delta(p^{\text{true}}(\mathbf{x}))
$$
where $\delta(\cdot)$ denotes the \emph{Dirac delta function}. For problems of practical interests, this limit is never reached, so it is necessary to deal with observations representing only a sparse subset of the possible attribute values. This paper presents a probabilistic methodology for handling such a situation. Our attention is limited to problems where $p(\mathbf{x})$ is an unknown $\mathtt{x}$-dimensional continuous function. Figure \ref{fig:Simulated_data} presents an unidimensional example of such a function where each of the $\mathtt{d}$ dots corresponds to a class observation, each associated with a different attribute value $x$. 

For each vector of attribute values $\mathbf{x}_{i}$, our knowledge of $p^{\text{true}}(\mathbf{x}_{j})$ conditional on an observation, $\hat{c}_{i}(\mathbf{x}_{i})$, is described by a Beta distribution as presented in Eq.\eqref{EQ:Beta_post}. In order to propagate the knowledge $f''\left(p(\mathbf{x}_{j})|\hat{c}_{i}(\mathbf{x}_{i})\right)$ to $f''\left(p(\mathbf{x}_{j}+\Delta\mathbf{x})|\hat{c}_{i}(\mathbf{x}_{i})\right)$, it is necessary to model their joint conditional \emph{pdf} as a \emph{random field}. $\mathcal{S}=\{\mathbf{x}_{j}\}_{j=1}^{\mathtt{s}}$ denotes the set of query attribute values where one is interested in predicting the joint probability $\mathbf{p}(\mathcal{S})$. Note that the set of query attribute values $\mathcal{S}$ contains the set of observed attribute values in $\mathcal{D}$, i.e. $\{\mathbf{x}:\mathbf{x}\in\mathcal{D}\}\subseteq\mathcal{S}$. For all $\mathbf{x}\in\mathcal{S}$, a joint conditional \emph{pdf} is formulated using Beta-distributed marginal \emph{pdf}s as presented in Eq.\eqref{EQ:Beta_post}, and a Gaussian copula function as presented by Der Kiureghian and Pei-Ling \cite{der1986structural}. This combination leads to the \emph{Nataf-Beta} probability distribution given by
\begin{equation}
\begin{array}{rcl}
f(\mathbf{p}(\mathcal{S})|\mathcal{D})&=&f\left(\underset{\mathbf{x}\in\mathcal{S}}{\cap}p\left(\mathbf{x}\right)|\mathcal{D}\right) \\[12pt]
&\triangleq&\text{NBeta}(\mathbf{p}(\mathcal{S});\mathbf{a}(\mathcal{S}),\mathbf{b}(\mathcal{S}),\mathbf{R}_{p})\\[4pt]
&=&\displaystyle\phi_{\mathtt{s}}(\mathbf{z}(\mathbf{p}(\mathcal{S}));\mathbf{R}_{p})\prod_{\mathbf{x}\in\mathcal{S}}\frac{\text{Beta}(p(\mathbf{x});a(\mathbf{x}),b(\mathbf{x}))}{\phi(z(p(\mathbf{x})))}\\[12pt]
\end{array}
\label{EQ:joint_pdf}
\end{equation}
where $Z\sim \phi(z)$ is a standard normal random variable for which the $\mathtt{s}$-dimensional joint \emph{pdf} is defined by $\phi_{\mathtt{s}}(\mathbf{z}(\mathbf{p}(\mathcal{S}));\mathbf{R}_{p})$. The transformation between the \emph{standard normal space} and the attribute space is given by
$$z(p(\mathbf{x}))=\Phi^{-1}[F(p(\mathbf{x}))]$$
where, $\Phi^{-1}[\cdot]$ is the inverse standard normal cumulative distribution function (\emph{cdf}). $\mathbf{R}_{p}$ is the correlation matrix defined in the standard normal space. The spatial correlation between $P(\mathbf{x}_{i})$ and $P(\mathbf{x}_{j})$ is governed by the Mahalanobis distance between $\mathbf{x}_{i}$ and $\mathbf{x}_{j}$. Accordingly, a \emph{Gaussian radial basis function kernel} \cite{murphy2012machine} describes $[\mathbf{R}_{p}]_{ij}$ so that
\begin{equation}
[\mathbf{R}_p]_{ij}=\exp\left(-\frac{1}{2}(\mathbf{x}_{i}-\mathbf{x}_{j})\text{diag}(\bm{l}_{p})^{-1} (\mathbf{x}_{i}-\mathbf{x}_{j})^{\intercal}\right)
\label{EQ:Rp}
\end{equation}
$\bm{l}_{p}=[l_{p,1},l_{p,2},\cdots,l_{p,\mathtt{x}}]^{\intercal}$ is a vector containing the \emph{length scale} parameter for each dimension of our attributes space. 

Following the hypothesis that $p^{\text{true}}(\mathbf{x})$ is a continuous function, our knowledge of it must be continuous as well. Here, our knowledge of $p^{\text{true}}(\mathcal{S})$ is fully described by $\mathbf{a}(\mathcal{S})$ and $\mathbf{b}(\mathcal{S})$, where each can take any positive real value and where it is assumed that $\mathbf{a}(\mathcal{S}){\perp\!\!\!\perp}\mathbf{b}(\mathcal{S})$. The prior knowledge of $\mathbf{a}^{\text{true}}(\mathcal{S})$ and $\mathbf{b}^{\text{true}}(\mathcal{S})$ is described by random fields, in this case, lognormal processes so that
\begin{equation}
\begin{array}{rcl}
\mathbf{A}'(\mathcal{S})\sim f'(\mathbf{a}(\mathcal{S}))&=&\ln\mathcal{N}(\mathbf{a}(\mathcal{S});\bm{\lambda}'_{a},\bm{\Sigma}'_{a})\\
\mathbf{B}'(\mathcal{S})\sim f'(\mathbf{b}(\mathcal{S}))&=&\ln\mathcal{N}(\mathbf{b}(\mathcal{S});\bm{\lambda}'_{b},\bm{\Sigma}'_{b})\\
\end{array}
\label{EQ:a_prior}
\end{equation}
where 
$$\begin{array}{rcl}
\bm{\Sigma_{a}}'&=&\text{diag}(\bm{\zeta}'_{a})\,\mathbf{R}_{a}\,\text{diag}(\bm{\zeta}'_{a})\\
\bm{\Sigma_{b}}'&=&\text{diag}(\bm{\zeta}'_{b})\,\mathbf{R}_{b}\,\text{diag}(\bm{\zeta}'_{b})\\
\end{array}$$
$\bm{\lambda}'=[\lambda'_{1},\lambda'_{2},\cdots,\lambda'_{\mathtt{s}}]^{\intercal}$ and $\bm{\zeta}'=[\zeta'_{1},\zeta'_{2},\cdots,\zeta'_{\mathtt{s}}]^{\intercal}$ are respectively the vectors of means and standard deviations of $a'(\mathbf{x})$ and $b'(\mathbf{x})$ taken in the log-space. Following Eq.\eqref{EQ:Rp} 
$$\begin{array}{rcl}
[\mathbf{R}_a]_{ij}&=&\exp\left(-\frac{1}{2}(\mathbf{x}_{i}-\mathbf{x}_{j})\text{diag}(\bm{l}_{a})^{-1} (\mathbf{x}_{i}-\mathbf{x}_{j})^{\intercal}\right)\\[8pt]
[\mathbf{R}_b]_{ij}&=&\exp\left(-\frac{1}{2}(\mathbf{x}_{i}-\mathbf{x}_{j})\text{diag}(\bm{l}_{b})^{-1} (\mathbf{x}_{i}-\mathbf{x}_{j})^{\intercal}\right)\\
\end{array}$$
Again, $\bm{l}_{a}$ and $\bm{l}_{b}$ are $\mathtt{x}$-dimensional vectors containing length scales corresponding to each dimension of the attribute space. Note that the lognormal processes in Eq.\eqref{EQ:a_prior} is only a transformation of a Gaussian Process for which an analytic formulation is already available \cite{williams2006gaussian}.

Following Eq.\eqref{EQ:ab_counts}, $\hat{a}(\mathbf{x})$ and $\hat{b}(\mathbf{x})$ denote the direct count of the number of observations in $\mathcal{D}$ of a given class for a specific set of attribute values so that
\begin{equation}
\begin{array}{rcl}
\hat{a}(\mathbf{x})&\triangleq&\#\{i:\{ \hat{c}_{i}(\mathbf{x})=1\}_{i=1}^{\mathtt{d}}\}\\[4pt]
\hat{b}(\mathbf{x})&\triangleq&\#\{i:\{ \hat{c}_{i}(\mathbf{x})=0\}_{i=1}^{\mathtt{d}}\}
\end{array}
\end{equation} 
$\hat{\mathbf{a}}(\mathcal{S})$ and $\hat{\mathbf{b}}(\mathcal{S})$ denote vectors containing the direct counts of the number of observations of a given class for each $\mathbf{x}\in\mathcal{S}$ so that
\begin{equation}
\begin{array}{rcl}
\hat{\mathbf{a}}(\mathcal{S})&=&[\hat{a}(\mathcal{S}_{1}),\hat{a}(\mathcal{S}_{2}),\cdots,\hat{a}(\mathcal{S}_{\mathtt{s}})]^{\intercal}\\[4pt]
\hat{\mathbf{b}}(\mathcal{S})&=&[\hat{b}(\mathcal{S}_{1}),\hat{b}(\mathcal{S}_{2}),\cdots,\hat{b}(\mathcal{S}_{\mathtt{s}})]^{\intercal}
\end{array}
\end{equation}

By knowing $\bm{l}_{a}$ and $\bm{l}_{b}$, it is possible to propagate the knowledge $\hat{a}(\mathbf{x}_{i}),\hat{b}(\mathbf{x}_{i})$, gained at a specific location $\mathbf{x}_{i}$ from a single observation $\hat{c}_{i}(\mathbf{x}_{i})\in\mathcal{D}$, to any other subset of attribute values, $a(\mathbf{x}),b(\mathbf{x}), \forall \mathbf{x}\in\mathcal{S}$. Given our assumption that the prior \emph{pdf} of $\mathbf{a}(\mathcal{S})$ and $\mathbf{b}(\mathcal{S})$ is respectively described by a log-normal process, their posterior \emph{pdf} is obtained using Gaussian posterior conditional  \cite{murphy2012machine} so that
\begin{equation}
\begin{array}{rcl}
f''\left(\underset{\mathbf{x}\in\mathcal{S}}{\cap} a(\mathbf{x})|\hat{a}(\mathbf{x}_{i})\right)&=&\ln\mathcal{N}(\mathbf{a}(\mathcal{S});\bm{\lambda}_{a}'',\bm{\Sigma}_{a}'')\\[12pt]
f''\left(\underset{\mathbf{x}\in\mathcal{S}}{\cap} b(\mathbf{x})|\hat{b}(\mathbf{x}_{i})\right)&=&\ln\mathcal{N}(\mathbf{b}(\mathcal{S});\bm{\lambda}_{b}'',\bm{\Sigma}_{b}'')
\end{array}
\label{EQ:ab_post}
\end{equation}   
where parameters $\bm{\lambda}'',\bm{\Sigma}''$ are obtained so that
\begin{equation}
\begin{array}{rcl}
\bm{\lambda}''_{a}&=&\bm{\lambda}_{a}'-[\bm{\Sigma}_{a}']_{:,i}\cdot \zeta_{a,i}^{'-1}\cdot\ln(\hat{a}(\mathbf{x}_{i}))\\[4pt]
\bm{\lambda}''_{b}&=&\bm{\lambda}_{b}'-[\bm{\Sigma}_{b}']_{:,i}\cdot\zeta_{b,i}^{'-1}\cdot\ln(\hat{b}(\mathbf{x}_{i}))
\end{array}
\end{equation}   
\begin{equation}
\begin{array}{rcl}
\bm{\Sigma}''_{a}&=&\bm{\Sigma}_{a}'-[\bm{\Sigma}_{a}']_{:,i}\cdot\zeta_{a,i}^{'-1}\cdot[\bm{\Sigma}_{a}']_{i,:}\\[4pt]
\bm{\Sigma}''_{b}&=&\bm{\Sigma}_{b}'-[\bm{\Sigma}_{b}']_{:,i}\cdot\zeta_{b,i}^{'-1}\cdot[\bm{\Sigma}_{b}']_{i,:}
\end{array}
\label{EQ:sigma_a_post}
\end{equation}
For the limit case where a length scale $l\to0$, no knowledge is transferred to attribute values other than the one directly measured; if $l\to\infty$, all attribute values share the same knowledge, i.e. $a(x_{i})=a(x_{j}),\forall i,j$, no matter how far they are from the attribute value observed. 

The respective posterior joint distribution of $\mathbf{a}(\mathcal{S})$ and $\mathbf{b}(\mathcal{S})$ conditional on the set of observations $\mathcal{D}$, is obtained by summing the posterior obtained in Eq.\eqref{EQ:ab_post} for each observation so  that
\begin{equation}\begin{array}{rcl}
\mathbf{A}''(\mathcal{S})\sim f''\left(\underset{\mathbf{x}\in\mathcal{S}}{\cap} a(\mathbf{x})|\mathcal{D}\right)&=&f''\left(\underset{\mathbf{x}\in\mathcal{S}}{\cap} a(\mathbf{x})|\underset{\mathbf{x}\in\mathcal{D}}{\cap}\hat{a}(\mathbf{x})\right)\\[12pt]
&=&\displaystyle\sum_{i=1}^{\mathtt{d}}f''\left(\underset{\mathbf{x}\in\mathcal{S}}{\cap} a(\mathbf{x})|\hat{a}(\mathbf{x}_{i})\right)\\[12pt]
\mathbf{B}''(\mathcal{S})\sim f''\left(\underset{\mathbf{x}\in\mathcal{S}}{\cap} b(\mathbf{x})|\mathcal{D}\right)&=&f''\left(\underset{\mathbf{x}\in\mathcal{S}}{\cap} b(\mathbf{x})|\underset{\mathbf{x}\in\mathcal{D}}{\cap}\hat{a}(\mathbf{x})\right)\\[12pt]
&=&\displaystyle\sum_{i=1}^{\mathtt{d}}f''\left(\underset{\mathbf{x}\in\mathcal{S}}{\cap} a(\mathbf{x})|\hat{b}(\mathbf{x}_{i})\right)
\end{array}
\label{EQ:ab_post}
\end{equation}
The joint posterior \emph{pdf} describing the probability of belonging to a particular class conditioned on observations is given by 
$$f''(\mathbf{p}(\mathcal{S})|\mathcal{D})=\text{NBeta}(\mathbf{p}(\mathcal{S});\mathbf{A}''(\mathcal{S}),\mathbf{B}''(\mathcal{S}),\mathbf{R}_{p})$$ 

In the above procedure, it is assumed that the length scale $\bm{l}_{a}$, $\bm{l}_{b}$, and $\bm{l}_p$ are known constants. In practical cases, it is necessary to learn what are their possible values from observations. Bayesian inference is employed to learn the posterior \emph{pdf} for length scale parameters, $\bm{l}_{a}$, $\bm{l}_{b}$, and $\bm{l}_{p}$ following 
$$
f''(\bm{l}_{a}, \bm{l}_{b},\bm{l}_{p}|\mathcal{D})=\frac{f'(\mathcal{D}|\bm{l}_{a}, \bm{l}_{b},\bm{l}_{p})f'(\bm{l}_{a}, \bm{l}_{b},\bm{l}_{p})}{f(\mathcal{D})}
$$
where, $f'(\bm{l}_{a}, \bm{l}_{b},\bm{l}_{p})$ is the joint \emph{pdf} describing prior knowledge, and $f(\mathcal{D})$ is the normalization constant. In order to derive an analytical formulation for the likelihood function $f'(\mathcal{D}|\bm{l}_{a}, \bm{l}_{b},\bm{l}_{p})$, it is necessary to notice that $\bm{l}_{a}$, $\bm{l}_{b}$ are \emph{hyper-parameters}, i.e., these are parameters of the prior knowledge. The joint posterior \emph{pdf} describing the probability of belonging to a particular class conditioned on observations is given by
$$f'(\mathbf{p}(\mathcal{S})|\mathcal{D})=\text{NBeta}(\mathbf{p}(\mathcal{S});\mathbf{A}'(\mathcal{S}),\mathbf{B}'(\mathcal{S}),\mathbf{R}_{p})\\
$$
where $\mathbf{A}'(\mathcal{S})$ and $\mathbf{B}'(\mathcal{S})$ represents only the knowledge that has been propagated from indirect observations so that
$$\begin{array}{rcl}
\mathbf{A}'(\mathcal{S})&=&\mathbf{A}''(\mathcal{S})-\hat{\mathbf{a}}(\mathcal{S})\\[4pt]
\mathbf{B}'(\mathcal{S})&=&\mathbf{B}''(\mathcal{S})-\hat{\mathbf{b}}(\mathcal{S})
\end{array}
$$
In the limit case where the length scales $\bm{l}_{a}=\bm{l}_{b}\to0$,
$$a'(\mathbf{x})=b'(\mathbf{x})\to0,\forall \mathbf{x}\in \mathcal{S}
$$
and in the other limit case where $\bm{l}_{a}=\bm{l}_{b}\to\infty$
$$\begin{array}{rcl}
a'(\mathbf{x})&\to&\sum (\hat{\mathbf{a}}(\mathcal{S}))-\hat{a}(\mathbf{x}) \\[8pt]
b'(\mathbf{x})&\to& \sum (\hat{\mathbf{b}}(\mathcal{S}))-\hat{b}(\mathbf{x})
\end{array}
$$
How much particular values of $\bm{l}_{a}$, $\bm{l}_{b}$ and $\bm{l}_{p}$ explain the set of observations $\mathcal{D}$ is quantified through the likelihood function 
\begin{equation}
\begin{array}{rcl}
f'(\mathcal{D}|\bm{l}_{a}, \bm{l}_{b},\bm{l}_{p})&=&\displaystyle\prod_{i=1}^{\mathtt{d}} \int\int\int p(\mathbf{x}_{i})^{a'(\mathbf{x}_{i})}(1-p(\mathbf{x}_{i}))^{b'(\mathbf{x}_{i})}\\[8pt]
&&\qquad \cdot \text{NBeta}(p(\mathbf{x}_{i});a'(\mathbf{x}_{i}),b'(\mathbf{x}_{i}),\mathbf{R}_{p})\\[8pt]
&&\qquad\cdot f(a'(\mathbf{x}_{i}))\cdot f(b'(\mathbf{x}_{i}))\partial a'(\mathbf{x}_{i})\partial b'(\mathbf{x}_{i})\partial p(\mathbf{x}_{i})
\end{array}
\label{EQ:likelihood}
\end{equation}
The likelihood function in Eq.\eqref{EQ:likelihood} has no closed-form analytic solution. An accurate approximation of the likelihood can be obtained using Monte-Carlo sampling techniques such as those presented by MacKay \cite{mackay1998introduction}. A computationally less demanding, but more crude approximation consists in using expected values $\mathbb{E}[\mathbf{A}'(\mathcal{S})]$ and $\mathbb{E}[\mathbf{B}'(\mathcal{S})]$ instead of the full probability densities. It reduces the likelihood function to
$$\begin{array}{rcl}
\hat{f}'(\mathcal{D}|\bm{l}_{a}, \bm{l}_{b},\bm{l}_{p})&=&\displaystyle\prod_{i=1}^{\mathtt{d}} \int p(\mathbf{x}_{i})^{\mathbb{E}[\mathbf{A}'(\mathcal{S})]}(1-p(\mathbf{x}_{i}))^{\mathbb{E}[\mathbf{B}'(\mathcal{S})]}\\[10pt]
&&\quad \cdot \text{NBeta}(p(\mathbf{x}_{i});\mathbb{E}[\mathbf{A}'(\mathcal{S})],\mathbb{E}[\mathbf{B}'(\mathcal{S})],\mathbf{R}_{p})\partial p(\mathbf{x}_{i})
\end{array}
$$
Unfortunately, this simplification is insufficient to lead to an analytically tractable solution; An analytically tractable solution is reached by making the simplifying assumption that the probabilities of classes for different sets of attribute values are independent so that $P(\mathbf{x}_{i}){\perp\!\!\!\perp}P(\mathbf{x}_{j}),\forall i\neq j$. In that case, the parameters $\bm{l}_{p}\equiv0$ and the likelihood reduces to 
\begin{equation}
\begin{array}{rcl}
\hat{f}'(\mathcal{D}|\bm{l}_{a}, \bm{l}_{b})&=&\displaystyle\prod_{i=1}^{\mathtt{d}} \left( \begin{array}{c}\mathtt{d}\\ \hat{a}(\mathbf{x}_{i}) \\\end{array}\right)\\
&&\quad\displaystyle\cdot\frac{\text{B}(\mathbb{E}[\mathbf{A}'(\mathcal{S})]+\hat{a}(\mathbf{x}_{i}),\mathbb{E}[\mathbf{B}'(\mathcal{S})]+\hat{b}(\mathbf{x}_{i}))}{\text{B}(\mathbb{E}[\mathbf{A}'(\mathcal{S})],\mathbb{E}[\mathbf{B}'(\mathcal{S})])}
\end{array}
\label{EQ:likelihood_approx_2}
\end{equation}
where $B(\cdot,\cdot)$ is the Beta function. 

The posterior predictive \emph{pdf} $\tilde{f}''(\mathbf{p}(\mathcal{S})|\mathcal{D})$ if given by
\begin{equation}
\tilde{f}''(\mathbf{p}(\mathcal{S})|\mathcal{D})=\int \mathbf{p}(\mathcal{S}) f''(\mathbf{p}(\mathcal{S})|\mathcal{D})\partial \mathbf{p}(\mathcal{S})
\label{EQ:post_predictive_1}
\end{equation}
Since no analytic formulation exist for the \emph{pdf} in Eq.\eqref{EQ:post_predictive_1}, it has to be evaluated numerically. By following the same simplifying assumptions as for Eq.\eqref{EQ:likelihood_approx_2}, an approximation of the posterior predictive \emph{pdf} is given by
\begin{equation}
\tilde{\hat{f}}''(\mathbf{p}(\mathcal{S})|\mathcal{D})=\frac{\mathbb{E}[\mathbf{A}''(\mathcal{S})]}{\mathbb{E}[\mathbf{A}''(\mathcal{S})]+\mathbb{E}[\mathbf{B}''(\mathcal{S})]}
\label{EQ:post_predictive_2}
\end{equation}

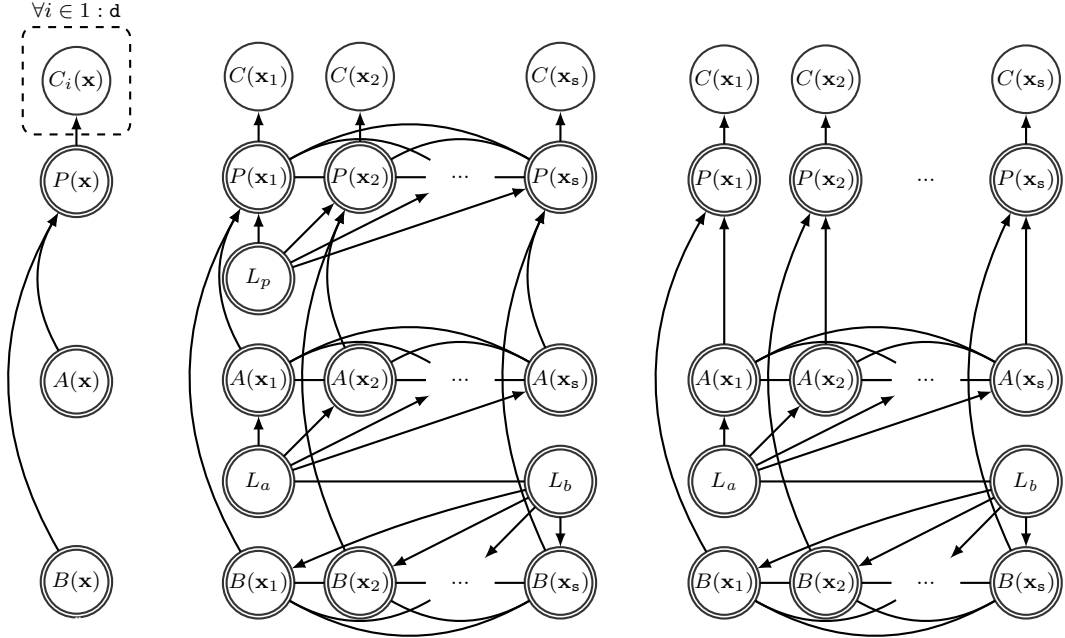
\begin{figure}[ht]
\hspace{-10mm}
  \begin{tabular}{ccc}
    \subfloat[Beta conjugate prior, Eq.\eqref{EQ:Beta_post}]{
\begin{tikzpicture}
\scriptsize
\node[cont] (Pxj) [label=center:$P(\mathbf{x})$] { };
\node[disc] (Cx1) [above=of Pxj, label=center:$C_{i}(\mathbf{x})$] { };
\node[dummy] (Lp) [below=of Pxj] { };
\node[cont] (Ax1) [below=of Lp, label=center:$A(\mathbf{x})$] { };
\node[dummy] (La) [below=of Ax1] { };

\node[cont] (Bx1) [below=of La, label=center:$B(\mathbf{x})$] { };
\node[dummy] (Lb) [above=of Bx1] { };

\path (Pxj) edge [connect] (Cx1) 
	(Ax1) edge [connect,bend left=30] (Pxj)
	
	(Bx1) edge [connect,bend left=30] (Pxj)
	(Bx1) edge [min distance=4mm, line, out=270, in=275,draw =white]  (Bx1)
         ;
\node[box, fit= (Cx1),label=above:$\forall i\in{1:\mathtt{d}}$] {};
\end{tikzpicture}}&
    \subfloat[Complete formulation of the Nataf-Beta Random Field Classifier as presented in Eq.\eqref{EQ:likelihood}]{
\begin{tikzpicture}
\scriptsize
\node[cont] (Px1) [label=center:$P(\mathbf{x}_{1})$] { };
\node[cont] (Px2) [right=of Px1, label=center:$P(\mathbf{x}_{2})$] { };
\node[dummy] (Pxn) [right=of Px2, label=center: ...] { };
\node[cont] (Pxns) [right=of Pxn, label=center:$P(\mathbf{x}_{\mathtt{s}})$] { };
\node[disc] (Cx1) [above=of Px1, label=center:$C(\mathbf{x}_{1})$] { };
\node[disc] (Cx2) [above=of Px2, label=center:$C(\mathbf{x}_{2})$] { };
\node[disc] (Cxns) [above=of Pxns, label=center:$C(\mathbf{x}_{\mathtt{s}})$] { };
\node[cont] (Lp) [below=of Px1, label=center:$L_{p}$] { };
\node[cont] (Ax1) [below=of Lp, label=center:$A(\mathbf{x}_{1})$] { };
\node[cont] (Ax2) [right=of Ax1, label=center:$A(\mathbf{x}_{2})$] { };
\node[dummy] (Axn) [right=of Ax2, label=center: ...] { };
\node[cont] (Axns) [right=of Axn, label=center:$A(\mathbf{x}_{\mathtt{s}})$] { };
\node[cont] (La) [below=of Ax1, label=center:$L_{a}$] { };

\node[cont] (Bx1) [below=of La, label=center:$B(\mathbf{x}_{1})$] { };
\node[cont] (Bx2) [right=of Bx1, label=center:$B(\mathbf{x}_{2})$] { };
\node[dummy] (Bxn) [right=of Bx2, label=center: ...] { };
\node[cont] (Bxns) [right=of Bxn, label=center:$B(\mathbf{x}_{\mathtt{s}})$] { };
\node[cont] (Lb) [above=of Bxns, label=center:$L_{b}$] { };

\path (Px1) edge [connect] (Cx1) 
	(Px2) edge [connect] (Cx2) 
	(Pxns) edge [connect] (Cxns)   
	(La) edge [connect] (Ax1)  
	(La) edge [connect] (Ax2)  
	(La) edge [connect] (Axns)  
	(La) edge [connect] (Axn)
	(Axn) edge [line] (Ax2)
	(Ax1) edge [line] (Ax2)
	(Ax2) edge [line] (Axn)
	(Axn) edge [line] (Axns)
	(Ax1) edge [connect,bend left=30] (Px1)
	(Ax2) edge [connect,bend left=25] (Px2)
	(Axns) edge [connect,bend left=25] (Pxns)

	(Ax1) edge [line,bend left=30] (Axn)
	(Ax2) edge [line,bend left=30] (Axns)
	(Ax1) edge [line,bend left=30] (Axns)
	(Lp) edge [connect] (Px1)  
	(Lp) edge [connect] (Px2)  
	(Lp) edge [connect] (Pxn)  
	(Lp) edge [connect] (Pxns)  
	
	(Lb) edge [connect,bend right=5] (Bx1)  
	(Lb) edge [connect] (Bx2)  
	(Lb) edge [connect] (Bxn)  
	(Lb) edge [connect] (Bxns)
	(Bx1) edge [connect,bend left=30] (Px1)
	(Bx2) edge [connect,bend left=25] (Px2)
	(Bxns) edge [connect,bend left=25] (Pxns)
	
	(Bx1) edge [line,bend right=30] (Bxn)
	(Bx1) edge [line,bend right=30] (Bxns)
	(Bx2) edge [line,bend right=30] (Bxns)
	
	(Bx1) edge [line] (Bx2)
	(Bx2) edge [line] (Bxn)
	(Bxn) edge [line] (Bxns)
	
	(Px1) edge [line] (Px2)
	(Px2) edge [line] (Pxn)
	(Pxn) edge [line] (Pxns)
	(Px2) edge [line,bend left=30] (Pxns)
	(Px1) edge [line,bend left=30] (Pxn)
	(Px1) edge [line,bend left=30] (Pxns)
	(La) edge [line] (Lb)
         ;
\end{tikzpicture}}&
        \subfloat[Simplified formulation of the Nataf-Beta Random Field Classifier as presented in Eq.\eqref{EQ:likelihood_approx_2}]{
\begin{tikzpicture}
\scriptsize
\node[cont] (Px1) [label=center:$P(\mathbf{x}_{1})$] { };
\node[cont] (Px2) [right=of Px1, label=center:$P(\mathbf{x}_{2})$] { };
\node[dummy] (Pxn) [right=of Px2, label=center: ...] { };
\node[cont] (Pxns) [right=of Pxn, label=center:$P(\mathbf{x}_{\mathtt{s}})$] { };
\node[disc] (Cx1) [above=of Px1, label=center:$C(\mathbf{x}_{1})$] { };
\node[disc] (Cx2) [above=of Px2, label=center:$C(\mathbf{x}_{2})$] { };
\node[disc] (Cxns) [above=of Pxns, label=center:$C(\mathbf{x}_{\mathtt{s}})$] { };
\node[dummy] (Lp) [below=of Px1] { };
\node[cont] (Ax1) [below=of Lp, label=center:$A(\mathbf{x}_{1})$] { };
\node[cont] (Ax2) [right=of Ax1, label=center:$A(\mathbf{x}_{2})$] { };
\node[dummy] (Axn) [right=of Ax2, label=center: ...] { };
\node[cont] (Axns) [right=of Axn, label=center:$A(\mathbf{x}_{\mathtt{s}})$] { };
\node[cont] (La) [below=of Ax1, label=center:$L_{\mathit{a}}$] { };
\node[cont] (Lb) [below=of Axns, label=center:$L_{\mathit{b}}$] { };

\node[cont] (Bx1) [below=of La, label=center:$B(\mathbf{x}_{1})$] { };
\node[cont] (Bx2) [right=of Bx1, label=center:$B(\mathbf{x}_{2})$] { };
\node[dummy] (Bxn) [right=of Bx2, label=center: ...] { };
\node[cont] (Bxns) [right=of Bxn, label=center:$B(\mathbf{x}_{\mathtt{s}})$] { };

\path (Px1) edge [connect] (Cx1) 
	(Px2) edge [connect] (Cx2) 
	(Pxns) edge [connect] (Cxns)   
	(La) edge [connect] (Ax1)  
	(La) edge [connect] (Ax2)  
	(La) edge [connect] (Axns)  
	(La) edge [connect] (Axn)
	(Axn) edge [line] (Ax2)
	(Ax1) edge [line] (Ax2)
	(Ax2) edge [line] (Axn)
	(Axn) edge [line] (Axns)
	(Ax1) edge [connect] (Px1)
	(Ax2) edge [connect] (Px2)
	(Axns) edge [connect] (Pxns)

	(Ax1) edge [line,bend left=30] (Axn)
	(Ax2) edge [line,bend left=30] (Axns)
	(Ax1) edge [line,bend left=30] (Axns)
	
	(Bx1) edge [connect,bend left=30] (Px1)
	(Bx2) edge [connect,bend left=25] (Px2)
	(Bxns) edge [connect,bend left=25] (Pxns)
	
	(Bx1) edge [line,bend right=30] (Bxn)
	(Bx1) edge [line,bend right=30] (Bxns)
	(Bx2) edge [line,bend right=30] (Bxns)
	
	(Bx1) edge [line] (Bx2)
	(Bx2) edge [line] (Bxn)
	(Bxn) edge [line] (Bxns)
	(Lb) edge [connect,bend right=5] (Bx1)  
	(Lb) edge [connect] (Bx2)  
	(Lb) edge [connect] (Bxn)  
	(Lb) edge [connect] (Bxns)
	(Lb) edge [line] (La)  
         ;
\end{tikzpicture}}
  \end{tabular}
  \caption{Graphical models describing (a) the Beta conjugate prior, (b) the complete and (c) simplified formulation for the Nataf-Beta Random Field Classifier.}
\label{fig:bayesian_network}
\end{figure}

Figure \ref{fig:bayesian_network} presents the graphical model \cite{murphy2012machine,Pearl:1988fk} for (a) the Beta conjugate prior, (b) the complete and (c) the simplified formulation for the Nataf-Beta Random Field Classifier. In the graphical models, circles represent random variables, arrows correspond to causal relations and links to bi-directional non-causal relations. Single-line nodes are discrete random variables; double-line nodes are continuous random variables. Note that for the graphical models in (b) and (c) there are one $C(\mathbf{x})$ per column of nodes; for (a), there are $\mathtt{d}$ $C_{i}(\mathbf{x})$. Also, in the special case where all observations are obtained for a same attribute value $\mathbf{x}_{i}=\mathbf{x}^{*}$, the Nataf-Beta Random Field Classifier collapses to the Beta-binomial conjugate prior. In such a special case, the graphical models in Figure \ref{fig:bayesian_network} (a), (b) and (c) are all equivalents.

\section{Empirical validation}\label{S:validation}
This section validates the performance of the Nataf-Beta Random Field Classifier using simulated and benchmark datasets. Both, rely on the same prior knowledge and use the same search algorithm to identify length scales. All results were obtained using the simplifying hypotheses taken in Eq.\eqref{EQ:likelihood_approx_2}, so that the length scale $\mathbf{l}_{p}\equiv0$. Also, the number of parameters is reduced by assuming that $\bm{l}_{a}\equiv\bm{l}_{b}$. 

The prior mean and variance of $A'(\mathbf{x})$ and  $B'(\mathbf{x})$ both tends to 0. For practical purposes  $\mathbb{E}[A'(\mathbf{x})]=\mathbb{E}[B'(\mathbf{x})]=10^{-10}$ and $\operatorname{var}[A'(\mathbf{x})]=\operatorname{var}[B'(\mathbf{x})]=10^{-20}$. 
For all following applications, the prior \emph{pdf}s for $\bm{l}_{a}$ and $\bm{l}_{b}$ is assumed to be uniform for $l\in\mathbb{R}^{+}$.

The maximum a posteriori (MAP) values for $\bm{l}_{a}$ and $\bm{l}_{b}$ are sought using a \emph{Newton-Raphson} gradient ascent method  \cite{murphy2012machine}. For all examples, only the MAP estimates are used. For every analyses, the start point for each length scale corresponds to the mean of the attribute values in the training set. The stopping criteria are either (1) if the difference in the mean log-likelihood ($\mathbb{E}[\ln\mathcal{L}]$) value over the 10 and 5 previous steps is less than $10^{-3}\times \mathbb{E}[\ln\mathcal{L}]$, (2) More than 100 iterations have been made, or (3) more than 2 hours is spent without reaching convergence.   

Two performance metrics are computed for characterizing \emph{classification accuracy}. In each case, the accuracy is quantified using a \emph{10-fold cross-validation procedure}, where the average of results obtained over each of the 10 test-sets are reported. The first metric is the correct classification rate ($\mathit{CCR}$) defined by
$$\mathit{CCR}\triangleq\frac{\mathit{TP}+\mathit{TN}}{\mathit{TP}+\mathit{TN}+\mathit{FP}+\mathit{FN}}
$$
where $\mathit{TP},\mathit{TN},\mathit{FP},\mathit{FN}$ respectively stands for \emph{true positive}, \emph{true negative}, \emph{false positive} and \emph{false negative}. In binary classification cases, an observation is deemed to belong to class $c$ (i.e. either a $\mathit{TP}$ or $\mathit{TN}$ instance) if the posterior predictive $\tilde{\hat{f}}''(\Pr(C(\mathbf{x})=c)|\mathcal{D})>0.5$. In the case where the classification problem has more than 2 classes, an observation is deemed to belong to class $c$ if the posterior predictive $\tilde{\hat{f}}''(\Pr(C(\mathbf{x})=c)|\mathcal{D})$ is the greatest among all other classes.

The second metric is the \emph{probability of correct classification} ($\mathit{PCC}$)
$$\mathit{PCC}\triangleq\prod_{i=1}^{\mathtt{d}} \tilde{\hat{f}}''(\Pr(C_{i}(\mathbf{x}_{i})=c^{*}_{i})|\mathcal{D})
$$
where $c^{*}$ denotes the most probable class so that
$$c^{*}_{i}=\underset{c_{i}}{\text{arg\,max}}~\tilde{\hat{f}}''(\Pr(C_{i}(\mathbf{x})=c_{i})|\mathcal{D})
$$
The probability of correct classification allows estimating the classification accuracy without using the data in the test set. If $\mathit{PCC}-\mathit{CCR}\to0$, it means the classification accuracy was predictable before observing any test data.

\subsection{Simulated data}
The first example consists in a binary classification problem where simulated data is generated using samples from the \emph{pdf} described by Eq.\eqref{EQ:joint_pdf}, for $\bm{l}_{a}=\bm{l}_{b}=\bm{l}_{p}=2$. The probability of observing an attribute value is uniformly distributed over the range (0,10). Figure \ref{fig:Simulated_data} presents the simulated $p^{\text{true}}(x)$, which corresponds to one realization of $f(\mathbf{p}(\mathcal{S})|\mathcal{D})$.

\begin{figure}[htbp]
\centering
  \begin{tabular}{c}
    \subfloat[10 observations]{
   	\includegraphics[width=80mm]{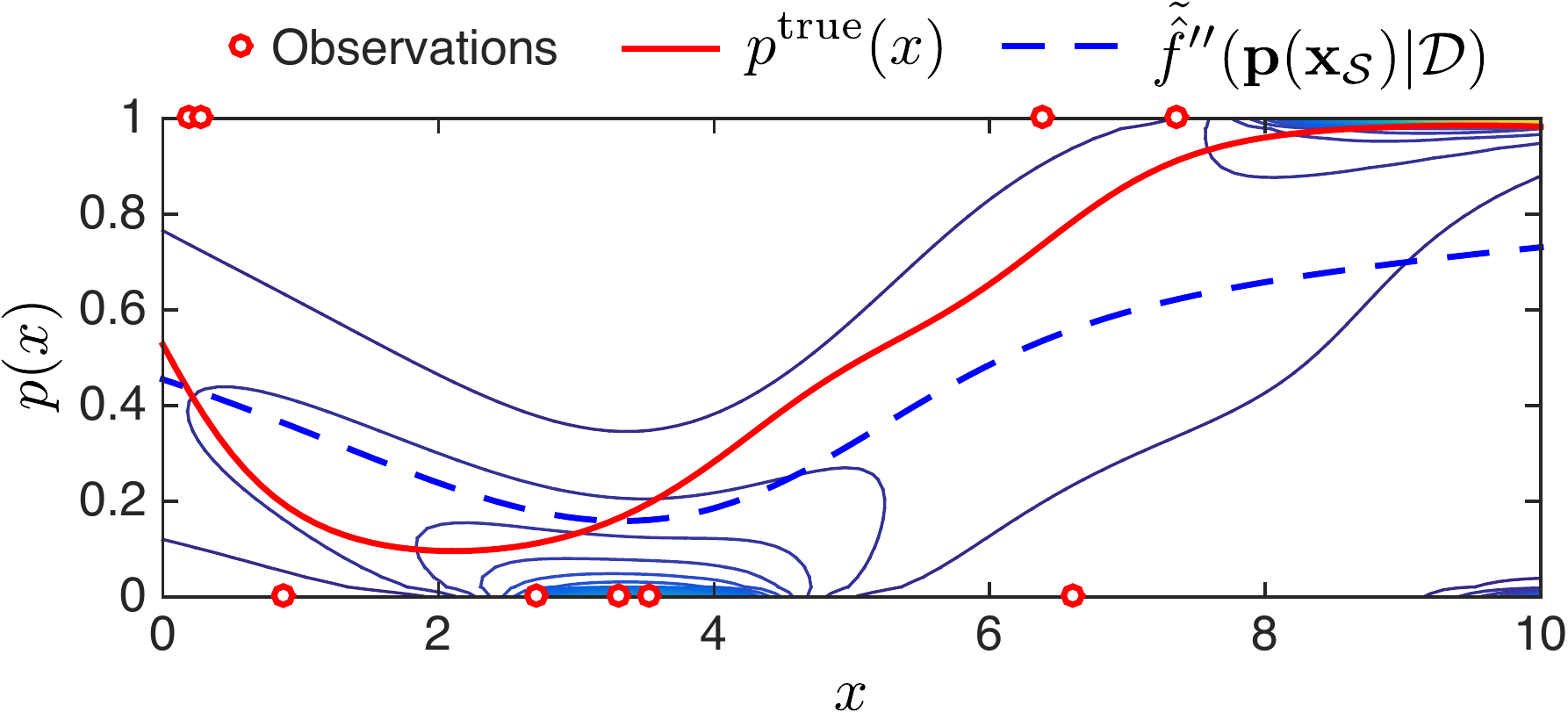}
    	\label{fig:subfig1_PNN}} \\
    \subfloat[100 observations]{
    	\includegraphics[width=80mm]{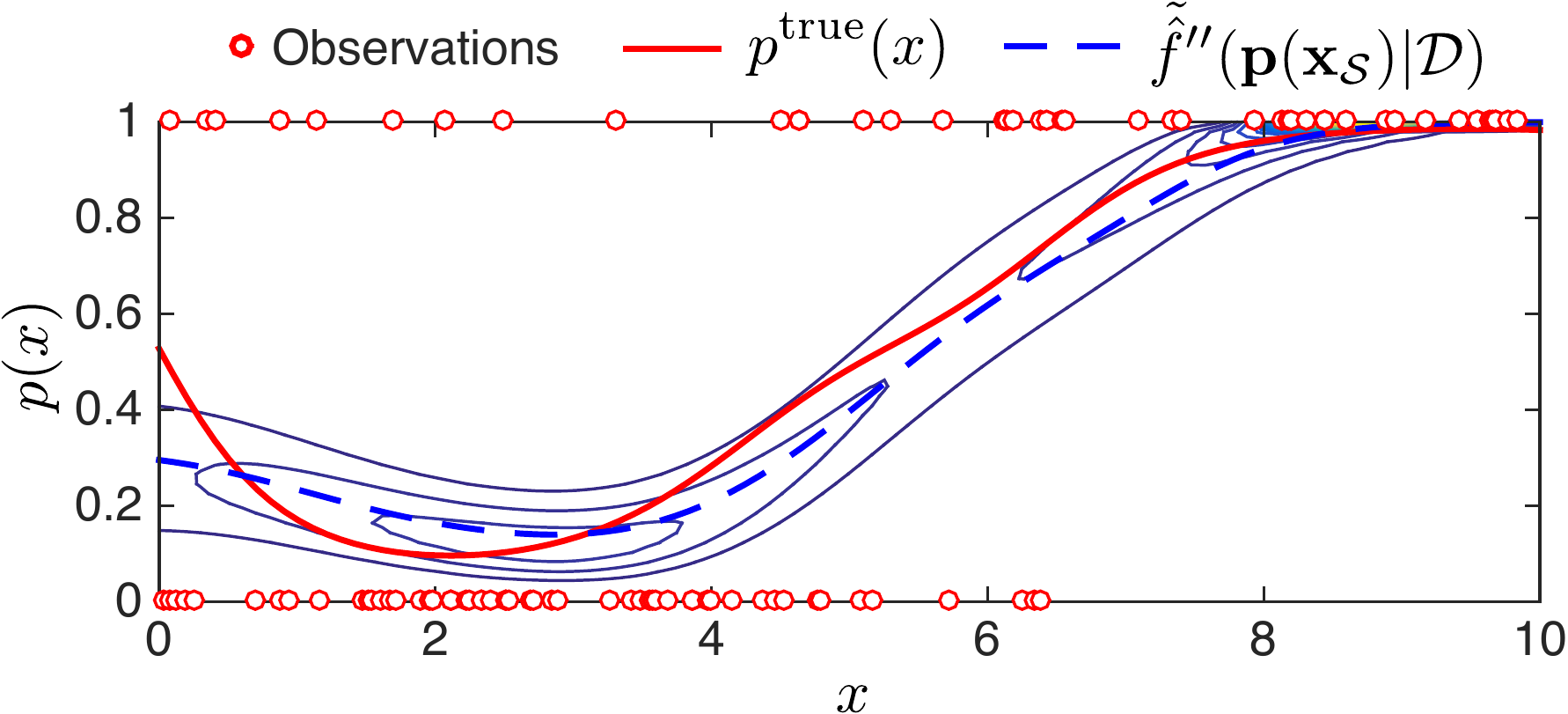}
    	\label{fig:subfig2_PNN}}\\
	 \subfloat[500 observations]{
    	\includegraphics[width=80mm]{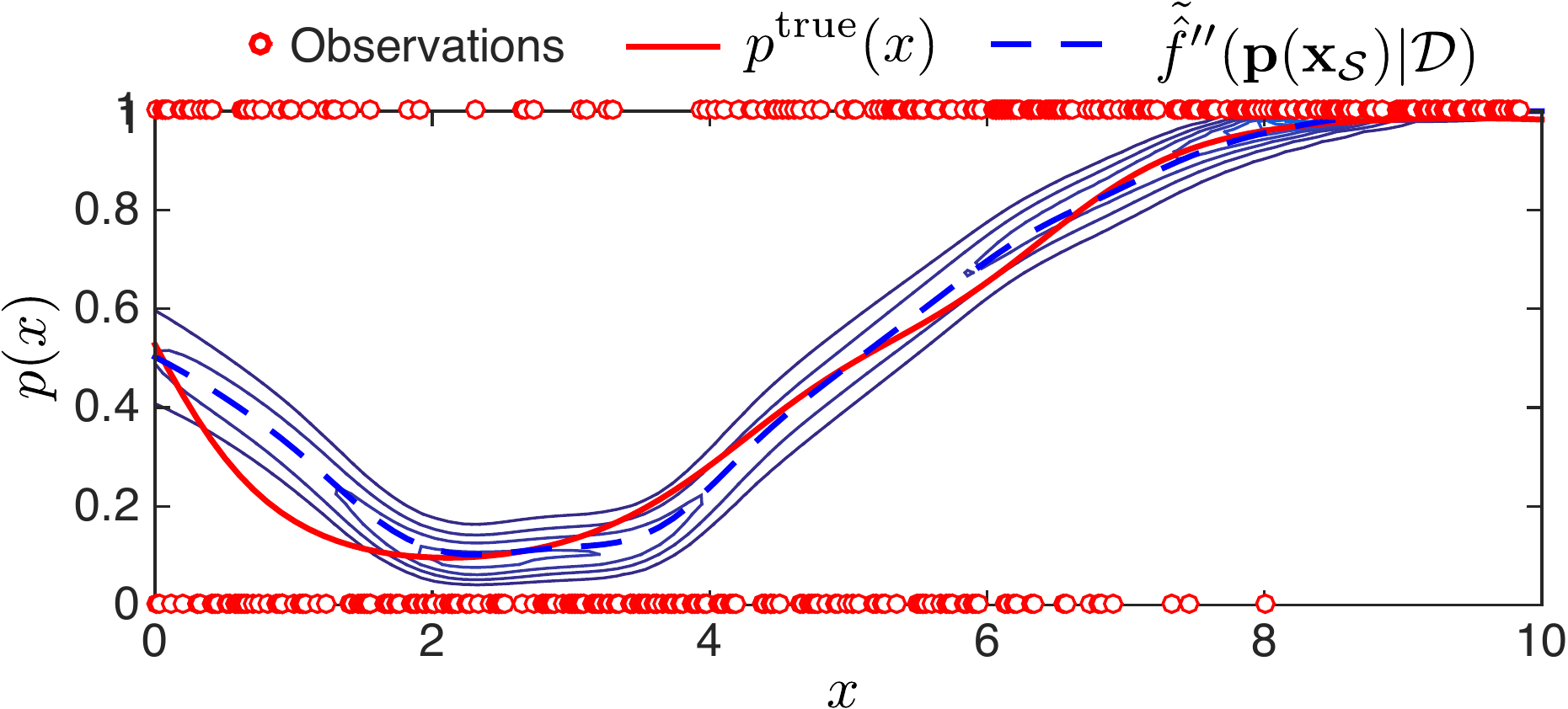}
    	\label{fig:subfig3_PNN}}
  \end{tabular} \hfill
  \caption{Comparison of the performance of the Nataf-Beta Random Field Classifier using (a) 10, (b) 100, and (c) 500 observations. Each plot shows the contours of the posterior \emph{pdf}, $\hat{f}''(\mathbf{p}(\mathcal{S})|\mathcal{D})$, the posterior predictive \emph{pdf}, $\tilde{\hat{f}}''(\mathbf{p}(\mathcal{S})|\mathcal{D})$ and the true values $p^{\text{true}}(x)$. Simulated observations are depicted by circles.}
\label{fig:PNN_nataf_beta}
\end{figure}

Table \ref{TAB:Results_nataf_beta} presents the classification accuracy obtained using 10, 100 and 500 simulated observations. Given that this is a simulated example, the true CCR and PCC accuracies are available. These results indicate that as the number of observations increases, the classification accuracy approach the true value.
\begin{table}[htbp]
  	\centering
  	\caption{Comparison of the \emph{classification accuracy} (CCR) and the \emph{probability of correct classification} (PCC) with their true values.}
	\begin{tabular}{rcc|cc|cc}
	\addlinespace
	\toprule
\#&\multicolumn{2}{c}{Accuracy - truth [\%]}&	\multicolumn{2}{c}{Accuracy - $\mathit{CCR}$  [\%]}&\multicolumn{2}{c}{Accuracy - $\mathit{PCC}$  [\%]}\\
observations &$CCR^{\text{true}}$& $PPC^{\text{true}}$& $\mathbb{E}[CCR]$& $\text{std}[CCR]$&$\mathbb{E}[PCC]$& $\text{std}[PCC]$ \\
	\cmidrule(lr){1-7}
	10& 70.2 & 67.9& 60.0&51.6 & 46.5 & 11.8 \\
	100& 81.0 & 73.1&81.0 & 12.9& 71.3 & 7.3 \\
	500& 79.0 & 71.2 &78.4&5.6&70.9&3.1\\
    	\bottomrule
    	\end{tabular}
  	\label{TAB:Results_nataf_beta}
\end{table}
Figure \ref{fig:PNN_nataf_beta} compares the performance of the Nataf-Beta Random Field Classifier using an increasing number of observations, (a) 10, (b) 100 and (c) 500. Each plot shows the contours of the posterior \emph{pdf}, $\hat{f}''(\mathbf{p}(\mathcal{S})|\mathcal{D})$, the posterior predictive \emph{pdf}, $\tilde{\hat{f}}''(\mathbf{p}(\mathcal{S})|\mathcal{D})$ and the true values $p^{\text{true}}(x)$. Again, as the number of observation increases, the contours of the posterior \emph{pdf} and the predictive posterior approach the true values $p^{\text{true}}(x)$.

\subsection{Benchmark datasets}\label{SS:Benchmark}
This section presents the accuracy of the Nataf-Beta Random Field Classifier for 6 real-continuous and real-integer attribute value benchmark datasets taken from UCI's Machine Learning Repository \cite{Bache:2015eu}. Table \ref{TAB:Description_UCI} provides target classification accuracy ranges for each dataset as reported by the following papers: \cite{shen2013prototype,Torre2004RT,keogh1999learning,polat2009new}. Note that in the references cited, no one methodology outperforms all others for all datasets. For the cases where a dataset contains missing data, missing values are replaced by the mean attribute value across the dataset.
\begin{table}[htb]
  	\centering
  	\caption{The accuracies reported are the ranges found in the following references: \cite{shen2013prototype,Torre2004RT,keogh1999learning,polat2009new}. Note that in the references cited, no one methodology outperforms all others for all datasets.} 
	\begin{tabular}{rcc}
	\addlinespace
	\toprule
&	Accuracy range  [\%] & Dataset\\
Dataset & $\mathbb{E}[CCR]$& Reference \\
	\cmidrule(lr){1-3}
	Iris &71.6--97.3&\cite{fisher1936use} \\
	Pima &55.3--75.4&\cite{smith1988using} \\
	Breast Cancer &89.7--97.0 &\cite{mangasarian1990cancer}\\
	Ionosphere &41.0--93.7&\cite{sigillito1989classification} \\
	Glass &38.1--95.5&\cite{evett1987rule}\\
	E.Coli &55.7--85.4&\cite{nakai1991expert}\\
    	\bottomrule
    	\end{tabular}
  	\label{TAB:Description_UCI}
\end{table}
 
The classification accuracies reached with the Nataf-Beta Random Field Classifier are presented in Table \ref{TAB:Results_uci_CV}. Although the classification accuracy does not in any case statistically outperform the best accuracy reported in the literature, it consistently ranks among the top tier. For the Iris, Cancer and Ionosphere datasets, the CCR and PPC accuracies are almost equals. This means that the predicted classification accuracy is itself accurate. For other datasets, the CCR and PPC values are consistent with each other, yet not as close. These results confirm that the proposed \emph{Nataf-Beta Random Field Classifier} is suited as a general purpose classification approach for real-continuous and real-integer attribute value problems.
\begin{table}[htb]
  	\centering
  	\caption{Nataf-Beta Random Field Classifier validation: Comparison of the \emph{classification accuracy} (CCR) and the \emph{probability of correct classification} (PCC) for UCI datasets. Results are averages obtained from the test sets of 10-fold cross-validation analyses.}
	\begin{tabular}{rcc|cc}
	\addlinespace
	\toprule
&	\multicolumn{2}{c}{Accuracy - $\mathit{CCR}$  [\%]}&\multicolumn{2}{c}{Accuracy - $\mathit{PCC}$  [\%]}\\
Dataset & $\mathbb{E}[CCR]$& $\text{std}[CCR]$&$\mathbb{E}[PCC]$& $\text{std}[PCC]$ \\
	\cmidrule(lr){1-5}
	Iris & 96.0&7.2 & 93.8&3.8\\
	Pima & 73.0& 6.0&65.6&4.4\\
	Breast Cancer &96.0&2.2&94.7& 2.2\\
	Ionosphere & 88.0& 4.6 & 87.1&3.7\\
	Glass &80.5 &6.5 &67.5 &8.3\\
	E.Coli &85.5 &6.2 &78.2 &5.7\\
    	\bottomrule
    	\end{tabular}
  	\label{TAB:Results_uci_CV}
\end{table}

\section{Comparison with Gaussian Process classification}\label{S:comparison}
Gaussian Process classification (GPC) is similar to the approach proposed in this paper because it models the spatial dependencies in the knowledge of the probability of a class using a Gaussian process. \cite{williams2006gaussian} describe Gaussian process classification as a ``\emph{natural generalization of the linear logistic regression model}''. The main idea is to define a Gaussian Process (GP) over the domain of attribute values and then map this GP to the probability space using a sigmoid-shaped transformation function such as the \emph{logit} or \emph{probit}. The parameters of that GP are identified using the calibration set $\mathcal{D}$. 

GPC can be seen as a \emph{bottom-up} approach; the start point is that we have the GP which provides a convenient analytic formulation for modeling spatial dependencies for real-valued outcomes. In order to be compatible with classification problems, these outcomes are transformed using a sigmoid function chosen to satisfy the requirement that the probability of a class, $\Pr(C(\mathbf{x})=1)$, must be defined over the interval $(0,1)$. In this case, the choice of the sigmoid-shaped transformation function and the parameter ($\bm{\theta}$) defining the Gaussian Process ($\text{GP}(\mathbf{y}(\mathcal{S});\bm{\theta}'')$) do not have a direct interpretation in relation with the classification problem. The posterior \emph{pdf} describing the probability of belonging to a given class is conceptually given by
$$f''(\mathbf{p}(\mathcal{S})|\mathcal{D})=\text{sigmoid}(\text{GP}(\mathbf{y}(\mathcal{S});\bm{\theta}''))$$
where $\mathbf{y}(\mathcal{S})$ is a set of real-valued outcomes obtained for each query point in $\mathcal{S}$.

Alternately, the \emph{Nataf-Beta Random Field Classifier} can be seen as a \emph{top-down} approach; the start point is that for a given a vector of attribute values $\mathbf{x}$, the classification problem is genuinely described by the Beta conjugate prior. The formulation presented in Eq.\eqref{EQ:joint_pdf} models the posterior \emph{pdf} describing the probability of belonging to a given class as a random field (i.e. a Nataf-Beta joint \emph{pdf}) for which marginal \emph{pdf}s are  Beta distributed. The posterior \emph{pdf} describing the probability of belonging to a given class is conceptually given by
$$f''(\mathbf{p}(\mathcal{S})|\mathcal{D})=\text{NBeta}(\mathbf{p}(\mathcal{S});\mathbf{A}''(\mathcal{S}),\mathbf{B}''(\mathcal{S}),\mathbf{R}_{p}'')$$
Given that $a(\mathbf{x})$ and $b(\mathbf{x})$ are positive real-valued number, and $\mathbf{A}''(\mathcal{S})$ and $\mathbf{B}''(\mathcal{S})$ are each modeled by random field, in this case, a log-normal process as described in Eq.\eqref{EQ:ab_post}.

Both approaches are providing a posterior joint \emph{pdf} describing the probability of belonging to a class for a set of query attribute values, $\mathcal{S}$. Therefore, both approaches are able to distinguish between a lack of knowledge, $\Pr(\text{class\,\#1})=\Pr(\text{class\,\#2})=0.5$, due to a lack of observations, and an intrinsically ambiguous class probabilities, i.e. $\Pr^{\text{true}}(\text{class\,\#1})=\Pr^{\text{true}}(\text{class\,\#2})=0.5$. The main difference between the \emph{Nataf-Beta Random Field Classifier} and \emph{Gaussian Process classification} is thus in the interpretation. The \emph{Nataf-Beta Random Field Classifier} has the same interpretation as the Beta-Bernouilli model and its formulation is directly issued from the binary classification problem; The \emph{Gaussian Process classification} is a powerful proxy capable of fitting complex functions $p^{\text{true}}(\mathbf{x})$, however, its formulation is not rooted in  classification problems.

\section{Discussion}\label{S:Discussion}
Results presented in \S\ref{SS:Benchmark} confirm that the \emph{Nataf-Beta Random Field Classifier} is suited as a general purpose classification approach for real-continuous and real-integer attribute value problems. Note that this performance is achieved despite making the following simplifying assumptions:
\begin{enumerate}
\item Only the MAP estimate for $a(\mathbf{x})$, $b(\mathbf{x})$ are employed.
\item Datasets containing more than two classes are analyzed as multiple 2-classes problems.
\item Missing data are replaced by the corresponding attribute mean value.
\item The identification of the length scale $\mathbf{l}_{p}$ is omitted.
\item The number of parameters is reduced by assuming that $\bm{l}_{a}\equiv\bm{l}_{b}$
\end{enumerate}
All these simplifications can be relaxed at the expense of computational resources.  Regarding the second simplification, a direct analysis of multi-classes problems is possible by using a Dirichlet \emph{pdf} instead of the Beta \emph{pdf} employed here. This extension is beyond the scope of this paper. 


It is important to note that as other classification methods, this one is not immune to the \emph{curse of dimensionality} \cite{murphy2012machine}. As the number of attributes ($\mathtt{x}$) increases, the sparsity of a dataset over the attribute $\mathtt{x}$-dimensional space increases exponentially. Guidance on how the accuracy of the proposed approach performs as a function of the sparsity of a dataset is beyond the scope of this paper.

\section{Conclusion}
The \emph{Nataf-Beta Random Field Classifier} is suited as a general purpose classification approach for real-continuous and real-integer attribute value problems. Although the classification accuracy does not statistically outperform the best accuracy reported in the literature, it consistently ranks among the top tier classification accuracies. The main strength of the approach resides in its formulation which extends the applicability of the Beta conjugate prior to classification problems. 

\section{Acknowledgements}
The author thanks the Swiss National Science Foundation for funding this research; Timothy Brathwaite for the numerous fruitful discussions related to machine learning; Professors Armen Der Kiureghian and Samer Madanat for kindly hosting him at UC Berkeley; The editor-in-chief of JMLR, Dr. Kevin Murphy, for providing useful comments such as pointing the similarity between the method proposed and Gaussian Process classification. 


\bibliographystyle{abbrv}
\bibliography{/Users/james_goulet/Dropbox/BibReader/Goulet_reference_librairy}


\end{document}